\documentclass{article}
\usepackage{ijcai19}

\usepackage{times}
\usepackage{soul}
\usepackage{url}
\usepackage[hidelinks]{hyperref}
\usepackage[small]{caption}
\usepackage{graphicx}
\usepackage{amsmath}
\usepackage{booktabs}
\usepackage{algorithm}
\usepackage{algorithmic}

\usepackage{array}
\usepackage{diagbox}
\usepackage{helvet}
\usepackage{courier}
\usepackage{url}
\usepackage{graphicx}
\usepackage{multirow}
\usepackage{subcaption}
\usepackage{array}
\usepackage{booktabs}
\usepackage{textcomp,booktabs}
\usepackage[usenames,dvipsnames]{color}
\usepackage{colortbl}
\definecolor{mygray}{gray}{.9}
\title{Neural Collective Entity Linking Based on Recurrent Random Walk \\ Network Learning}
\author{
Mengge Xue$^{1,2,3}$\and
Weiming Cai$^3$\and
Jinsong Su$^3$\thanks{Corresponding author}\and
Linfeng Song$^4$\and
Yubin Ge$^4$\and\\
Yubao Liu$^4$\And
Bin Wang$^5$
\affiliations
Institute of Information Engineering, Chinese Academy of Sciences$^1$\\
School of Cyber Security, University of Chinese Academy of Sciences$^2$\\
Xiamen University$^3$\\
Rochester University$^4$\\
Xiaomi AI Lab, Xiaomi Inc., Beijing, China$^5$\\
\emails
xuemengge@iie.ac.cn,\quad
caiweiming@stu.xmu.edu.cn,\quad
jssu@xmu.edu.cn
}

\begin{document}

\maketitle

\begin{abstract}
  Benefiting from the excellent ability of neural networks on learning semantic representations, existing studies for entity linking (EL) have resorted to neural networks to exploit both the local mention-to-entity compatibility and the global interdependence between different EL decisions for target entity disambiguation. However, most neural collective EL methods depend entirely upon neural networks to automatically model the semantic dependencies between different EL decisions, which lack of the guidance from external knowledge. In this paper, we propose a novel end-to-end neural network with recurrent random-walk layers for collective EL, which introduces external knowledge to model the semantic interdependence between different EL decisions. Specifically, we first establish a model based on local context features, and then stack random-walk layers to reinforce the evidence for related EL decisions into high-probability decisions, where the semantic interdependence between candidate entities is mainly induced from an external knowledge base. Finally, a semantic regularizer that preserves the collective EL decisions consistency is incorporated into the conventional objective function, so that the external knowledge base can be fully exploited in collective EL decisions. Experimental results and in-depth analysis on various datasets show that our model achieves better performance than other state-of-the-art models. Our code and data are released at \url{https://github.com/DeepLearnXMU/RRWEL}.
\end{abstract}

\section{Introduction}

Entity linking (EL) is a task to link the name mentions in a document to their referent entities within a knowledge base (KB). The great significance of the research on EL can not be neglected due to the solid foundation it helps to build for multiple natural language processing tasks, such as information extraction \cite{Ji:TAC2016}, semantic search \cite{Blanco:WSDM2015} and so on. Nevertheless, this task is non-trivial because mentions are usually ambiguous, and the inherent disambiguation between mentions and their referent entities still maintains EL as a challenging task.

The previous studies on EL are based on the statistical models, which mainly focus on artificially defined discriminative features,
%which mainly focus on the utilization of artificially defined discriminative features between mentions and its target entities,
%
which mainly focus on the utilization of artificially defined discriminative features between mentions and its target entities,
 such as contextual information, topic information or entities type etc.
 %such as contextual information, topic information etc.
 %\cite{Xin:AAAI2004,Adafre:LinkKDD2005,Bunescu:EACL2006,Cucerzan:EMNLP2007,Mihalcea:CIKM2007,Fader:WIKIAI2009,Cassidy:TAC2010,Dredze:COLING2010,Gaffney:COLING2010,Tan:COLING2010,Zheng:NAACL2010,Ji:ACL2011,Chen:EMNLP2011,Mendes:I-Semantics2011,Chisholm:TACL2015,Lazic:TACL2015,Shen:TKDE2015,Yamada:CONLL2016}.
\cite{Chisholm:TACL2015,Shen:TKDE2015}. However,
%one drawback of these approaches lies in that
 these models resolve mentions independently relying on textual context information from the surrounding words while ignoring the interdependence between different EL decisions. In order to address this problem, researches then devoted themselves to implementing collective decisions in EL, which encourages re entities of all mentions in a document to be semantically coherent %\cite{Medelyan:AAAI2008,Milne:ACM2008,Kulkarni:SIGKDD2009,Ferragina:CIKM2010,Hoffart:EMNLP2011,Ratinov:ACL2011,Shen:WWW2012,Cheng:EMNLP2013,He:EMNLP2013,Alhelbawy:ACL2014,Huang:ACL2014,Ganea:WWW2015,Pershina:NAACL2015,Globerson:ACL2016,Guo:SW2018,Phan:Arxiv2018}.
\cite{Hoffart:EMNLP2011,Ganea:WWW2015,Globerson:ACL2016,Guo:SW2018}.

Recently,
%with the rapid development of deep learning,
 the studies of EL have evolved from the conventional statistical models into the neural network (NN) based models thanks to their outstanding advantages in encoding semantics and dealing with data sparsity. Similarly, the studies of NN-based EL models have also experienced the development progress from models of independent decisions \cite{He:EMNLP2013,Francislandau:NAACL2016,Yamada:CONLL2016} to those of collective decisions \cite{Ganea:EMNLP2017,Cao:COLING2018,Le:ACL2018,kolitsas2018end-to-end}. For example,
%Huu et al., \shortcite{Huu:COLING2016} used convolutional neural network (CNN) to model the local features and simultaneously used recurrent neural network to model global features for EL.
 %Huu et al., \shortcite{Huu:COLING2016} used recurrent neural network to model global features for EL.
 %Ganea and Hofmann \shortcite{Ganea:EMNLP2017} applied a neural attention mechanism over local context windows combined with CRF to model the topical coherence.
 Ganea et al., \shortcite{Ganea:EMNLP2017} and Le et al., \shortcite{Le:ACL2018} solved the global training problem via truncated fitting loopy belief propagation (LBP).
 Cao et al., \shortcite{Cao:COLING2018} applied Graph Convolutional Network (GCN) to integrate global coherence information for EL.
 %Le et al., \shortcite{Le:ACL2018} treat relations between textual mentions in a document as latent variables in our neural entity-linking model.
 %Up to now, how to effectively leverage the interdependence between EL decisions to refine NN-based EL models has become a hot area of research in this field.
 However, in these works, they completely depend on NNs to automatically model the semantic dependencies between different EL decisions, while little attention has been paid on the guidance from an external KB.

In this paper, we propose a novel end-to-end neural collective model named Recurrent Random Walk based EL (RRWEL) which not only implements collective EL decisions but also leverages an external KB to model the semantic dependencies between candidate entities. Given a document,
%we first follow Han et al. \cite{Han:SIGIR2011} to collect candidate entities for all mentions and
%represent them with a referent graph, where each mention is linked to its all candidate entities, and two candidate entities of different mentions are connected.
%Based on the referent graph, we
 we first utilize some local features to measure the conditional probability of a mention referring to a specific entity, and then introduce random-walk layers to model the interdependence between different EL decisions, where the evidence for related EL decisions are computed based on the semantic relatedness between their corresponding knowledge pages, and can be fine-tuned during training. Finally, a semantic regularizer, which aims to preserve the collective EL decision consistency, is added to supplement the conventional EL loss. In that case, both the local semantic correspondence between mentions and entities and the global interdependence between different EL decisions can be fully exploited in our model to conduct collective EL decisions.
%follow Francis-Landau et al., \shortcite{Francislandau:NAACL2016} to
%apply CNNs to learn the semantic representations of each mention and its referent entity,
%which can be used to infer semantic distances of different levels,
%and then adopt a sparse linear model to convert the cosine similarities based on CNN-based representations
 %apply CNNs combined with a sparse linear model to measure the conditional probability of a mention referring to a specific entity,
%where the evidences for related EL decisions are reinforced into high-probability decisions
%according to the semantic interdependence between their candidate entities.
\begin{figure*}[!htbp]
\centering
\includegraphics[width=1.0\textwidth]{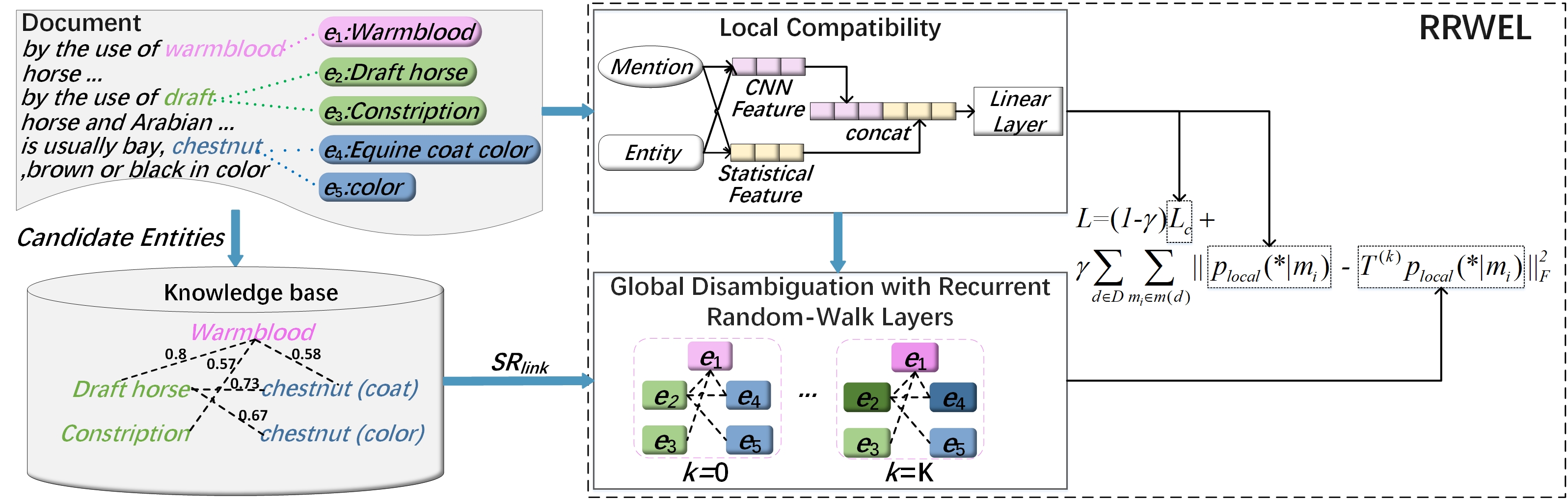}
\caption{
\label{Ourmodel_Framework_Fig}
The architecture of our proposed RRWEL model.
}
\end{figure*}

The main contributions of our work can be summarized as follows: (1) Through the in-depth analysis, we point out the drawbacks of dominant NN-based EL models and then dedicate to the study of employing KB based global interdependence between EL decisions for collective entity disambiguation; (2) We propose a novel end-to-end graph-based NN model which incorporates the external KB to collectively infer the referent entities of all mentions in the same document. To the best of our knowledge, this is the first NN-based random walk method for collective EL to explore external KB when modeling semantic interdependence between different EL decisions; (3) We evaluate the performance of our proposed model on many large-scale public datasets. The extensive experiments and results show that our model can outperform several state-of-the-art models.

\section{The Proposed Model}

In this section, we give a detailed description to our proposed model. As shown in Fig. \ref{Ourmodel_Framework_Fig}, our model mainly includes two components: one is used to model local mention-to-entity compatibility, the other leverages external knowledge to capture global entity-to-entity semantic interdependence.

More specifically, we first employ CNNs to learn the semantic representations of each mention and its all candidate entities, which are further fed into a linear model layer with other feature vectors to measure their local semantic compatibility. Then, we introduce random-walk layers to model the global interdependence between different EL decisions, which enable the EL evidence for related mentions to be collectively reinforced into high-probability EL decisions. In particular, during model training, in order to effectively exploit both the local mention-to-entity compatibility and the global interdependence between entities for collective ELs, we augment the conventional EL loss with a semantic regularizer.
%a consistency penalty term
%implying the collective EL decision consistency.

To clearly describe our model, we first introduce some annotations which are used in our model. Let $d$ be the input document and $m(d)$ = $\{m_1, m_2, ..., m_{N_{md}}\}$ be the set of mentions occurring in $d$, for each mention $m_i$, its candidates set $\Gamma(m_i)$ is identified. In addition,  for all mentions $\{m_i\}$, we collect all their candidate entities to form the candidate entity set $e(d)$ = $\{e_1, e_2, ..., e_{N_{ed}}\}$, where each entity $e_j$ has a corresponding KB page $p_j$. Using EL models, we expect to correctly map each mention to its corresponding entity page of the referenced KB.

\subsection{Local Mention-to-Entity Compatibility}

Following Francis-Landau et al., \shortcite{Francislandau:NAACL2016},
we represent the mention $m_i$ using the semantic information at three kinds of granularities: $s_i$, $c_i$, $d_i$, where $s_i$ is the \emph{surface string} of $m_i$, $c_i$ is the \emph{immediate context} within a predefined window of $m_i$ and $d_i$ is the \emph{entire document} contains $m_i$. For the candidate entity page $p_j$, we use its \emph{title} $t_j$ and \emph{body content} $b_j$ to represent it,
$p_j$ = ($t_j, b_j$).
%For convenience, we also denote $p_i$ = ($t_i$ , $b_i$) for the correct entity pages.
%Again, $t_{ij}$, $b_{ij}$, $t_i$ and $b_i$ are also sequences of words.

%With these annotations,
To correctly implement EL decisions, we first calculate the relevance score $\phi(m_i, p_j)$
%the conditional score $p(p_{ij}|m_i)$
%for each candidate entity page $p_{ij}$ of $m_i$,
%and finally link $m_i$ to the candidate entity page with the maximal score.
%Specifically, we compute $\phi$($m_i$, $p_{ij}$) in the following steps:
 for the candidate entity page $p_j$, and then normalize relevance scores to obtain the conditional probability $p_{local}(p_j|m_i)$:
\begin{align}
p_{local}(p_j|m_i) = \frac{\phi(m_i, p_j)}{\sum_{j'\in \Gamma(m_i)} \phi(m_i, p_{j'})},
\end{align}
where $\phi(m_i, p_j)$ is defined as follows:
\begin{align}
\phi(m_i, p_j) = \sigma(W_{\text{local}}[F_{\text{sf}}(m_i, p_j);F_{\text{cnn}}(m_i, p_j)]),
\end{align}
\iffalse
\begin{align}
&\phi(m_i, p_j) \notag \\
%&\phi(p_{ij}|m_i) \notag \\
=&\sigma(\phi_{\text{sparse}}(m_i, p_j) + \phi_{\text{CNN}}(m_i, p_j)) \\
=&\sigma(W_{\text{local}}[F_{\text{sf}}(m_i, p_j):F_{\text{cnn}}(m_i, p_j)]), \notag
\end{align}
\fi
where $\sigma(*)$ is the sigmoid function,
$F_{\text{sf}}(m_i, p_j)$ and $F_{\text{cnn}}(m_i, p_j)$ are two feature vectors that will be concatenated to obtain the local feature vectors, and $W_{\text{local}}$ is the weight for local feature vectors.
 In the following, we give detail descriptions of $F_{\text{sf}}(m_i, p_j)$ and $F_{\text{cnn}}(m_i, p_j)$.

1). $F_{\text{sf}}(m_i, p_j)$ denotes the statistical local feature vector proposed by Cao et al. \shortcite{Cao:COLING2018}. It contains various linguistic properties and statistics, including candidates' prior probability, string based similarities, compatibility similarities and the embeddings of contexts, all of which have been proven to be effective for EL. Due to the space limitation, we omit the detail descriptions of $F_{\text{sf}}(m_i, p_j)$. Please refer to \cite{Cao:COLING2018} for the details.

2). $F_{\text{cnn}}(m_i, p_j)$ combines the cosine similarities between the representation vectors of $m_i$ and $p_j$ at multiple granularities.
Formally, it is defined as follows:
\begin{align}
F_{\text{cnn}}(m_i, p_j)=[
&\cos(\bar{s}_i, \bar{t}_j), \cos(\bar{c}_i, \bar{t}_j),
\cos(\bar{d}_i, \bar{t}_j), \notag \\ &\cos(\bar{s}_i, \bar{b}_j),
\cos(\bar{c}_i, \bar{b}_j), \cos(\bar{d}_i, \bar{b}_j)],
\end{align}
where $\bar{s}_i$, $\bar{c}_i$, $\bar{d}_i$, $\bar{t}_j$, and $\bar{b}_j$ denote the distributed representations of $s_i$, $c_i$, $d_i$, $t_j$, and $b_j$, respectively.

%\textbf{First},
 In order to obtain the representations for the context word sequences of mentions and entities, we follow Francis-Landau et al., \shortcite{Francislandau:NAACL2016} and adopt CNNs to transform a context word sequence $x$$\in$\{$s_i$, $c_i$, $d_i$, $t_j$, $b_j$\} into the distributed representations, where each word of $x$ is represented by a real-valued, $h$-dimensional vector. The convolution operation is first performed on $w$, where $w$=[$w_1;w_2;...;w_N$]$\in$$R^{h\times N}$ is a matrix representing $x$. Here $N$ is the word number of $x$. Then we transform the produced hidden vector sequences by a non-linear function $G(*)$ and use \emph{Avg} function as pooling. Specifically, we employ one window size $l$ to parameterize the convolution operation, where the window size $l$ corresponds to a convolution matrix $M_l \in R^{v\times l\times h}$ of dimensionality $v$. Hence, we obtain the distributed representation $\bar{x}$ for $x$ as $\frac{1}{N-l}\sum^{N-l+1}_{i=1} G(M_l w_{i:(i+l-1)})$. Please note that we use the same set of convolution parameters for each type of text granularity in the source document $d$ as well as for the target entities.

At this stage, we are able to calculate $p_{local}(p_j|m_i)$ which provides abundant information of local mention-to-entity compatibility. The next step is to propagate this information based on the global interdependence between EL decisions.

\subsection{Global Interdependence Between EL Decisions}

Inspired by the random walk based EL \cite{Han:SIGIR2011} and the successful adaptation of random walk propagation to NN \cite{Zhao:IJCAI2017}, we introduce recurrent random-walk layers to propagate EL evidence for the purpose of effectively capturing the global interdependence between different EL decisions for collective entity predictions.
%In this aspect,
%there have been many methods proposed to calculate the semantic interdependence between entities.
%Due to its simplicity and effectiveness,
%here we directly adopt the method proposed by Milne and Witten \shortcite{Milne:CIKM2008} to compute the %semantic interdependence between entities as follows:

%\begin{align}
%SR(a,b)=1-\frac{\log(\max(|A|,|B|))-\log(|A|\bigcap|B|)}{\log(|W|)-\log(\min(|A|,|B|))},
%\end{align}
%where $a$ and $b$ are the two entities of interest,
%$A$ and $B$ are the sets of all entities that link to $a$ and $b$ in Wikipedia respectively,
%and $W$ denotes the entire Wikipedia.

To implement the propagation of EL evidence based on random walk, we first need to define a transition matrix $T$ between candidate entities,
where $T_{ij}$ is the evidence propagation ratio from $e_j$ to $e_i$. Intuitively, the more semantically related two entities are, the more evidence should be propagated between their EL decisions. Therefore, we calculate the semantic relevance between $e_i$ and $e_j$, and then normalize these relevance scores by entity to generate $T$:
%calculate their semantic interdependence:
\begin{align}
T(i,j)&=\frac{p(e_i\rightarrow e_j)}{\sum_{j'\in N_{e_i}}p(e_i\rightarrow e_{j'})},\\
p(e_i\rightarrow e_j)&=\text{SR}_{\text{link}}(p_i,p_j)+\text{SR}_{\text{semantic}}(p_i,p_j),
\end{align}
where $N_{e_i}$ is the set of neighbor entities of entity $e_i$. To be specific, for two considered entity pages $p_i= (t_i, b_i)$ and $p_j= (t_j, b_j)$,
we use hyperlinks to compute the semantic relevance score $\text{SR}_{\text{link}}(p_i,p_j)$:
\begin{align}
\text{SR}_{\text{link}}(p_i,p_j)=1-\frac{log(max(|I|,|J|))-log(|I\cap J|)}{log(|W|)-log(min(|I|,|J|))},
\end{align}
where $I$ and $J$ are the sets of all entities that link to $p_i$ and $p_j$ in KB respectively,
and $W$ is the entire KB.
Meanwhile,
we obtain their cosine similarity $\text{SR}_{\text{semantic}}(p_i,p_j)$ based on their CNN semantic representations.
Considering that the semantic relevance score between two candidate entities relies on the relative distance of their corresponding mentions, we supplement the conventional entity page $p_i=(t_i,b_i)$ with the position embedding $pos_i$ of its entity.
Formally,
$\text{SR}_{\text{semantic}}(p_i,p_j)$ is defined as follows:
\begin{align}
\text{SR}_{\text{semantic}}(p_i,p_j)= \cos([\bar{t}_i;\bar{e}_i]+pos_i,[\bar{t}_j;\bar{e}_j]+pos_j).
\end{align}
Here we follow Vaswani et al., \shortcite{vaswani:NIPS2017} to define the embedding $pos_i$ of mention $m_i$.
%Here $F^{ee}_{\text{CNN}}(*,*)$ denotes the concatenated vector of cosine similarities, with weight $W^{ee}_{\text{CNN}}$.
%Here $\text{SR}_{\text{link}}(p_i,p_j)$ leverages external knowledge to calculate semantic interdependence based on hyperlinks proposed by Milne and Witten \shortcite{Han:CIKM2009},
%where $p_i$ and $p_j$ are the two entities of interest,
%$I$ and $J$ are the sets of all entities that link to $p_i$ and $p_j$ in KB respectively, and $W$ is the entire KB.
%$\text{SR}_{\text{semantic}}(p_i,p_j)$ denotes the cosine similarity of concatenated vectors,
%where we follow vaswani et al., \shortcite{vaswani:NIPS2017} to define the embedding $pos_i$ of mention $m_i$.

With the transition matrix $T$, we perform the evidence propagation of EL decisions in the recurrent random-walk layers: (See the recurrent random-walk layers of Fig. \ref{Ourmodel_Framework_Fig})

\begin{align}
%\gamma^{k+1} &= T\cdot \gamma^{k} \notag \\
%             &= T^{(k)}\cdot \gamma^{0} \notag \\
%             &= T^{(k)}\cdot p(p_{j}|m_i),
p_{rw}^{(k+1)}(*|m_i) &= (1-\lambda)T\cdot p_{rw}^{(k)}(*|m_i) +\lambda p_{rw}^{(0)}(*|m_i)  \notag \\
                      &= (1-\lambda)T^{(k)}\cdot p_{rw}^{(0)}(*|m_i) +\lambda p_{rw}^{(0)}(*|m_i)  \notag \\
                      &= (1-\lambda)T^{(k)}\cdot p_{local}(*|m_i)+\lambda p_{local}(*|m_i),
\end{align}
%with $p^{(k)}(p_{j}|m_i)$ being the evidence vector through $k$ rounds of propagations.
%at the $k$-th iteration.
%with $\gamma^{k}$ being the evidence vector at the $k$-th iteration.
%Particularly, $\gamma^{0}$ = $p^{(k)}(p_{j}|m_i)$ which only exploits local mention-to-entity compatibility.
with $p_{rw}^{(k)}(*|m_i)$ being the predicted entity distribution of $m_i$ at the $k$-th iteration. Please note that $p_{rw}^{(0)}(*|m_i)$ = $p_{local}(*|m_i)$ which only exploits local mention-to-entity compatibility. Obviously, by introducing $K$ random-walk layers, we can easily propagate evidence for $K$ times based on random walk propagation.

\subsection{Model Training}

Aiming at combining the global interdependence between EL decisions with the local mention-to-entity context compatibility, we propose to not only minimize the common compatibility-based EL loss but also preserve the high-order EL consistency during model training. In this way, we can significantly improve our model by embedding the global interdependence between different EL decisions into the training of CNNs for modeling mention-to-entity compatibility.

%Let the vector $\hat{y}^{(k)}_j$ denote the predicted entity distribution of the $j$-th mention at time $k$,
%then its entity prediction contagion at time $k$+1 can be modeled
%by performing one step random walk process,
%of which is modeled as follows:
%$\hat{y}^{(k+1)}_j = \textbf{w}\hat{y}^{(k)}_j = \textbf{w}^{(k)}\hat{y}_j$,
%where the vector $\hat{y}_j$ is the predicted entity distribution of the $j$-th mention at time 0.
%Thus, the $k$-th order EL consistency of the $j$-th mention is preserved by minimizing the error
%$\| \hat{y}_i - \sum_{w_{i,j}>0,w_{i,j}\in\textbf{w}} w_{i,j}\hat{y}_j \|^2_F$,
%where the weight $w^{(k)}_{i,j}$ is from the $k$-th order transition matrix $\textbf{w}^{(k)}$,
%and $\|\cdot\|^2_F$ is the Frobenius norm.

The intuition behind our implementation lies in the convergence propriety of random walk process based on Markov chain \cite{gilks1995markov}.
Specifically, after multiple rounds of EL evidence propagation, the predicted entity distributions for mentions will tend to converge. If the global interdependence between different EL decisions has been well embedded into our model, we can expect that $p_{local}(*|m_i)$ $\approx$ $p_{rw}^{(K)}(*|m_i)$ $=$ $T^{(K)}\cdot p_{local}(*|m_i)$. To do this, we minimize the following error to preserve the the $K$-th order EL consistency of $m_i$, formulated as $\| p_{local}(*|m_i) - T^{(K)}\cdot p_{local}(*|m_i) \|^2_F$,
%where the weight $w^{(k)}_{i,j}$ is from the $k$-th order transition matrix $\textbf{w}^{(k)}$,
%and $\|\cdot\|^2_F$ is the Frobenius norm.
where $\|\cdot\|^2_F$ is the Frobenius norm.
\begin{figure}[t]
\centering
\includegraphics[scale=0.5]{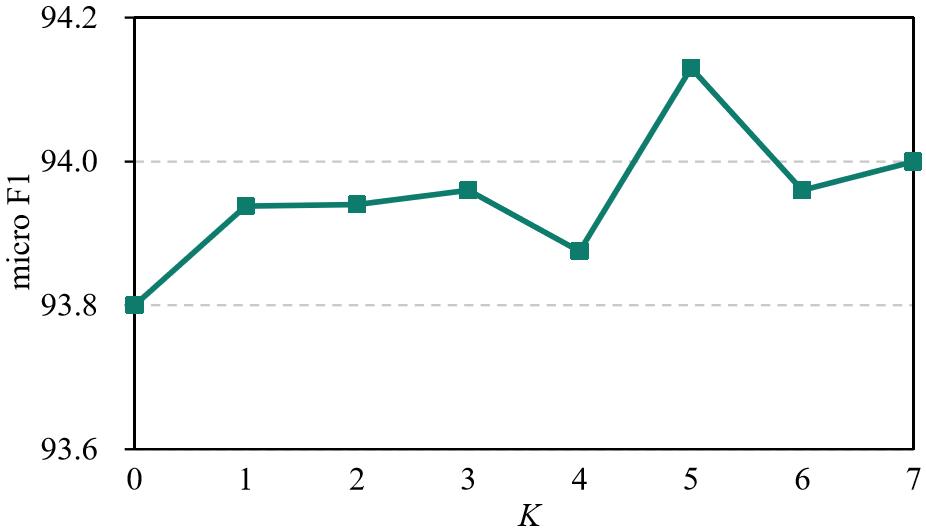}
\caption{
\label{KEffect-Fig}
Experimental results on the validation set AIDA-A using different numbers of random-walk layers.
}
\end{figure}

Finally, the recurrent random walk network learning for collective EL
can be mathematically formulated as
\begin{align}\label{Objective_Function1}
L&=(1-\gamma) \cdot L_c +  \notag \\
 &\gamma \cdot \sum_{d\in D}\sum_{m_i\in m(d)}\| p_{local}(*|m_i) - T^{(K)}\cdot p_{local}(*|m_i) \|^2_F,
\end{align}
where $D$ denotes the training corpus,
$K$ is the number of random-walk layers,
and the coefficient $\gamma$ is used to balance the preference between the two terms.
Specifically, the first term $L_c$ denotes the cross-entropy loss between predicted and ground truth $y^g$, which is defined as follows:
\begin{align}
L_c=-\sum_{d\in D}\sum_{m_{i}\in m(d)}\sum_{e_j\in \Gamma(m_i)}y^g_jlogp_{local}(e_j|m_i),
\end{align}

To train the proposed model, we denote all the model parameters by $\theta$. Therefore, the objective function in our learning process is given by
\begin{align}\label{Objective_Function2}
\min L(\theta) = L + \alpha \|\theta\|^2,
\end{align}
where $\alpha$ is the trade-off parameter between the training loss $L$ and regularizer $\|\theta\|^2$. To optimize this objective function, we employ the stochastic gradient descent (SGD) with diagonal variant of \emph{AdaGrad} in \cite{Duchi:JMLR2011}. Particularly, when training the model, we first use $L_c$ to pre-train the model and then use $L$ to fine-tune it.

\section{Experiment}

\subsection{Setup}\label{Experiment_Setup_Section}

\textbf{Datasets}.
We validated our proposed model on six different benchmark datasets used by previous studies:
\begin{itemize}
\setlength{\itemsep}{5pt}
\item \textbf{AIDA-CONLL}: This dataset is a manually annotated EL dataset \cite{Hoffart:EMNLP2011}. It consists of AIDA-train for training, AIDA-A for validation and AIDA-B for testing and there are totally 946, 216, 231 documents respectively.

\item \textbf{MSNBC, AQUAINT, ACE2004}: These datasets are cleaned and updated by Guo and Barbosa \shortcite{Guo:SW2018}, which contain 20, 50 and 36 documents respectively.

\item \textbf{WNED-WIKI (WIKI), WNED-CWEB (CWEB)}: These datasets are automatically extracted from ClueWeb and Wikipedia in \cite{Guo:SW2018,Gabrilovich:2013} and are relatively large with 320 documents each.
\end{itemize}
 In our experiments, we investigated the system performance with AIDA-train for training and AIDA-A for validation, and then tested on AIDA-B and  other datasets. Following previous works \cite{Ganea:EMNLP2017,Cao:COLING2018,Le:ACL2018}, we only considered mentions that have candidate entities in the referenced KB.

\textbf{Contrast Models}.
%We compared our RRWEL and RRWEL(local) to the baseline systems including three local and four types of global models:
%We compared RRWEL to the baseline systems including three local and four types of global models:
We compared our proposed RRWEL model to the following models:

\begin{table}[t]
\renewcommand\arraystretch{1.1}
    \centering
    %\small

    \begin{tabular}{|p{1cm}<{\centering}|p{0.75cm}<{\centering}|p{0.75cm}<{\centering}|p{0.75cm}<{\centering}|p{0.75cm}<{\centering}|p{0.75cm}<{\centering}|}
    \hline
        %\textbf{$\gamma$} & \multicolumn{5}{c|}{\textbf{$\lambda$} \\
        \diagbox{$\gamma$}{$\lambda$} & \textbf{0.1} & \textbf{0.3} &  \textbf{0.5} & \textbf{0.7} &  \textbf{0.9} \\
        \hline
        \textbf{0.1} & 93.64 & 93.9 & \textbf{94.13} & 93.92 & 94.00\\
        %\hline
        \textbf{0.3} & 92.64 & 93.49 & 93.47 & 93.22 & 93.3\\
        %\hline
        \textbf{0.5} & 90.94 & 91.51 & 91.51 & 91.13 & 92.58\\
        %\hline
        \textbf{0.7} & 85.35 & 85.67 & 88.15 & 87.43 & 89.03\\
        %\hline
        \textbf{0.9} & 69.72 & 69.48 & 69.03 & 68.57 & 68.91\\
        \hline
    \end{tabular}
    \caption{\label{gamaEffect}
    Experimental results on the validation set AIDA-A using different trade-off parameters.}
\end{table}
\begin{table}[t]
\renewcommand\arraystretch{1.1}
    \centering
    %\small
    \begin{tabular}{|p{5cm}<{\centering}|p{1.5cm}<{\centering}|}

        \hline
        \textbf{Model} & \textbf{AIDA-B} \\
        \hline
        %\emph{Local models}                                     &       \\
        Yamada et al., \shortcite{Yamada:CONLL2016}                     & 91.5  \\
        Francis-Landau et al., \shortcite{Francislandau:NAACL2016}       & 86.9  \\
        %RRWEL(local)                                              & \textbf{91.45}\\
        \hline
        \hline
        %\emph{Global models}                                    &       \\
        Ganea and Hofmann \shortcite{Ganea:EMNLP2017}                       & 92.22 \\
        Cao et al., \shortcite{Cao:COLING2018}                          & 87.2 \\
        Guo and Barbosa \shortcite{Guo:SW2018}                              & 89.0  \\
        Kolitsas et al., \shortcite{kolitsas2018end-to-end}                 &89.1   \\
        Le and Titov \shortcite{Le:ACL2018}                               & \textbf{93.07} \\
        \hline
        %RRWEL                                                    & 92.86\\
        RRWEL                                                    & 92.36\\
        \hline
    \end{tabular}
    \caption{\label{in-domain}
    Micro F1 scores for AIDA-B (in-domain) test set.
}
\end{table}
\begin{table*}[!ht]
\renewcommand\arraystretch{1.1}
    \centering
    \small

    \begin{tabular}{|p{5cm}<{\centering}|p{1.5cm}<{\centering}|p{1.5cm}<{\centering}|p{1.5cm}<{\centering}|p{1.5cm}<{\centering}|p{1.5cm}<{\centering}|p{1.5cm}<{\centering}|}
    \hline
        \textbf{Model} & \textbf{MSNBC} & \textbf{AQUAINT} &  \textbf{ACE2004} & \textbf{CWEB} &  \textbf{WIKI} & \textbf{Avg} \\
        \hline

        Hoffart et al., \shortcite{Hoffart:EMNLP2011}         &  79  &  56  &  80  & 58.6 &  63  & 67.32 \\
        Han et al., \shortcite{Han:SIGIR2011}                 &  88  &  79  &  73  &  61  &  78  & 75.8  \\
        Cheng and Roth \shortcite{Cheng:EMNLP2013}            &  90  &  90  &  86  & 67.5 & 73.4 & 81.38 \\
        Ganea and Hofmann \shortcite{Ganea:EMNLP2017}         & 93.7 & 88.5 & 88.5 & 77.9 & 77.5 & 85.22 \\
        Le and Titov \shortcite{Le:ACL2018}                   & 93.9 & 88.3 & 89.9 & 77.5 & 78.0 & 85.51 \\
        Guo and Barbosa \shortcite{Guo:SW2018}                &  92  &  87  &  88  &  77  & 84.5 & 85.7  \\
        %Cao et al., \shortcite{Cao:COLING2018}               &  &  &  &  &   &  \\
        \hline
        %RRWEL(local)                 & 91.36 & \textbf{91.79} & 91.06 & 79.29 & 84.16 & 87.532 \\
        %RRWEL                       & 92.82 & 90.62 & \textbf{91.49} & \textbf{80.03} & \textbf{84.52} & \textbf{87.9} \\
        RRWEL                       & \textbf{94.43} & \textbf{91.94} & \textbf{90.64} & \textbf{79.65} & \textbf{85.47} & \textbf{88.43} \\
        \hline
    \end{tabular}
    \caption{\label{out-domain}
    Performance (\textbf{Micro F1}) of various EL models on different datasets (out-domain). Particularly, we highlight the highest score in bold for each set.}

\end{table*}

\begin{table*}[!htbp]
	\centering
    \small
	\begin{tabular}{|p{4cm}<{\centering}|p{3cm}<{\centering}|p{2cm}<{\centering}|p{3cm}<{\centering}|p{2cm}<{\centering}|}
		\hline
        \multicolumn{5}{|p{14cm}|}{\emph{   \qquad The ideal W¨¹rttemberger stands around  high, and is usually bay, \textcolor{red}{chestnut}, \textcolor{blue}{brown}, or black in \textcolor{green}{color}}} \\
        \hline
        Ground-truth Entity & \multicolumn{2}{c|}{RRWEL($K$$=$$0$)} & \multicolumn{2}{c|}{RRWEL($K$$=$$5$)} \\
		\hline
		\emph{\textcolor{red}{chestnut (color)}} & \emph{\textcolor{red}{chestnut (color)}} & 0.996 & \emph{\textcolor{red}{chestnut (color)}} & 0.764 \\
		\hline
        \multirow{2}{*}{\emph{\textcolor{blue}{Bay (horse)}}}
        & \emph{\textcolor{blue}{Bay (horse)}} & 0.031 & \emph{\textcolor{blue}{Bay (horse)}} & 0.513 \\
        & \emph{\textcolor{blue}{Brown}} & 0.949 & \emph{\textcolor{blue}{Brown}} & 0.447 \\
        \hline
        \multirow{2}{*}{\emph{\textcolor{green}{Equine coat color}}}
        & \emph{\textcolor{green}{Equine coat color}} & 0.273 & \emph{\textcolor{green}{Equine coat color}} & 0.501 \\
        & \emph{\textcolor{green}{Color}} & 0.483 & \emph{\textcolor{green}{Color}} & 0.376 \\
        \hline
	\end{tabular}
    \caption{\label{effect of random walk}
    %Qualitative analysis of recurrent random-walk layers
    An example of the predicted entity distribution using different numbers of the recurrent random-walk layers}
\end{table*}
\begin{table*}[!htbp]
\renewcommand\arraystretch{1.1}
    \centering
    \small

    \begin{tabular}{|p{4cm}<{\centering}|p{1.5cm}<{\centering}|p{1.5cm}<{\centering}|p{1.5cm}<{\centering}|p{1.5cm}<{\centering}|p{1.5cm}<{\centering}|p{1.5cm}<{\centering}|}
    \hline
        \textbf{Model} & \textbf{AIDA-B} & \textbf{MSNBC} & \textbf{AQUAINT} &  \textbf{ACE2004} & \textbf{CWEB} &  \textbf{WIKI} \\
        \hline
        RRWEL(NN learning)                           & 91.32 & 93.14 & 90.90 & 90.21 & 78.45 & 84.37  \\
        RRWEL($\text{SR}_{\text{link}})$             & 92.00 & 93.90  & 91.35 & \textbf{91.06} & \textbf{80.14} & 84.95 \\
        RRWEL                                        & \textbf{92.36} & \textbf{94.43} & \textbf{91.94} & 90.64 & 79.65 & \textbf{85.47} \\
        \hline
    \end{tabular}
    \caption{\label{effect of external KB}
    Performance (\textbf{Micro F1}) under the effect of different transition matrixs on test datasets. Specifically, ``RRWEL(NN learning)" indicates that the transition matrix is automatically modeled by NN, while ``RRWEL($\text{SR}_{\text{link}})$" illustrates that we only use hyperlink information to initialize the transition matrix.}
\end{table*}
\iffalse
\begin{table*}[!htbp]
\renewcommand\arraystretch{1.1}
    \centering
    \small

    \begin{tabular}{|p{4cm}<{\centering}|p{1.5cm}<{\centering}|p{1.5cm}<{\centering}|p{1.5cm}<{\centering}|p{1.5cm}<{\centering}|p{1.5cm}<{\centering}|p{1.5cm}<{\centering}|}
    \hline
        \textbf{Model} & \textbf{AIDA-B} & \textbf{MSNBC} & \textbf{AQUAINT} &  \textbf{ACE2004} & \textbf{CWEB} &  \textbf{WIKI} \\
        \hline
        RRWEL(NN learning)                           & 91.16 & 93.30 & 90.91 & 89.79 & 78.45 & 83.84  \\
        RRWEL($\text{SR}_{\text{link}})$             & \textbf{92.35} & 93.90  & 91.50 & 90.64 & 79.79 & 84.82 \\
        RRWEL                                        & 91.35 & \textbf{93.95} & \textbf{92.33} & \textbf{90.64} & \textbf{79.91} & \textbf{85.04} \\
        \hline
    \end{tabular}
    \caption{\label{effect of external KB}
    Performance (\textbf{Micro F1}) under the effect of different transition matrixs on test datasets. Specifically, ``RRWEL(NN learning)" indicates that the transition matrix is automatically modeled by NN, while ``RRWEL($\text{SR}_{\text{link}})$" illustrates that we only use hyperlink information to initialize the transition matrix.}
\end{table*}
\fi

\begin{itemize}
\setlength{\itemsep}{5pt}
\item \textbf{\cite{Hoffart:EMNLP2011}} where iterative greedy method is employed to compute a subgraph with maximum density for EL.
\item \textbf{\cite{Han:SIGIR2011}} introduces random walk algorithm to implement collective entity disambiguation.
\item \textbf{\cite{Cheng:EMNLP2013}} utilizes integer linear programming to solve global entity linking.
\item \textbf{\cite{Francislandau:NAACL2016}}  applies CNN to learn the semantic representations of each mention and its all candidate entities.
\item \textbf{\cite{Yamada:CONLL2016}} proposes a novel embedding method specifically designed for NED which jointly maps words and entities into a same continuous vector space.
\item \textbf{\cite{Ganea:EMNLP2017}} solves the global training problem via truncated fitting LBP.
\item \textbf{\cite{Le:ACL2018}} encodes the relations between mentions as latent variables and applies LBP for global training problem.
\item \textbf{\cite{Guo:SW2018}} proposes a greedy, global NED algorithm which utilizes the mutual information between probability distributions induced from random walk propagation on the disambiguation graph.
%a sound information-theoretic notion of semantic interdependence derived from random walks on disambiguation graphs.
\item \textbf{\cite{Cao:COLING2018}} applies Graph Convolutional Network to integrate both local contextual features and global coherence information for EL.
\iffalse
\item \textbf{Local models}:Francislandau\cite{Francislandau:NAACL2016} also applies CNN to learn the semantic representations of each mention and its all candidate entities, Chisholm\cite{Chisholm:TACL2015} and NTEE\cite{Yamada:TACL2017} achieves  great performance using web links and joint embeddings of words and entities respectively.

\item \textbf{Iterative Greedy Model}:AIDA\cite{Hoffart:EMNLP2011} utilizes iterative greedy method to compute a subgraph with maximum density for EL.

\item \textbf{Loopy Belief Propagation}:Globerson\cite{Globerson:ACL2016} introduce LBP techniques for collective inference, G\&H\cite{Ganea:EMNLP2017} and L\&T\cite{Le:ACL2018} solves the global training problem via truncated fitting LBP.

\item \textbf{Random Walk}:Han11\cite{Han:SIGIR2011} and WNED\cite{Guo:SW2018} utilize random walks to solve collective entity disambiguation.

\item \textbf{Graph Convolutional Network}:NCEL\cite{Cao:COLING2018} applies Graph Convolutional Network to integrate both local contextual features and global coherence information for EL.
\fi
\end{itemize}

\textbf{Model Details}.
We used the latest English Wikipedia dump$\footnote{https://dumps.wikimedia.org/enwiki/20181101/enwiki-20181101-pages-articles-multistream.xml.bz2}$ as our referenced KB.
However, please note that our proposed model can be easily applied to other KBs. To employ CNN to learn the distributed representations of inputs,
we used 64 filters with the window size 3 for the convolution operation and the non-linear transformation function \emph{ReLU}.
Meanwhile, to learn the context representations of input mentions and target entities, we directly followed Francis-Landau \shortcite{Francislandau:NAACL2016} to utilize the window size 10 for context, and only extracted the first 100 words in the documents for mentions and entities. Besides, the standard \emph{Word2Vec} toolkit \cite{Mikolov:NIPS2013} was used with vector dimension size 300, window context size 21, negative sample number 10 and iteration number 10 to pre-train word embeddings on Wikipedia and then we fine-tuned them during model training. Particularly, following Ganea and Hofmann \shortcite{Ganea:EMNLP2017}, we kept top 7 candidates for each mention based on their prior probabilities, and while propagating the evidence, we just kept top 4 candidates for each mention $m_i$ according to $p_{local}(*|m_i)$. Besides, we set $\alpha$$=$$1$$e$$-$$5$. Finally, we adopted standard F1 score at Mention level (Micro) as measurement.

\subsection{Effects of $K$ and Trade-off Parameters}

There are three crucial parameters in our approach, which are the number $K$ of random-walk layers, the trade-off parameter $\lambda$ for restart and the trade-off parameter $\gamma$ for loss function. We tried different hyper-parameters according to the performance of our model on the validation set.

To investigate the effect of the trade-off parameter $\lambda$ and $\gamma$, we kept $K$$=$$5$ and then tried different combinations of these two parameters with $\lambda$ from 0.1 to 0.9 and $\gamma$ from 0.1 to 0.9. The results are shown in Table \ref{gamaEffect}. We observed that our method achieves the best performance when the trade-off parameters $\lambda$ and $\gamma$ are set to 0.5 and 0.1 respectively. Consequently, all the following experiments only consider the proposed model with $\lambda$$=$$0.5$ and $\gamma$$=$$0.1$.

Besides, we varied the value of $K$ from 0 to 7 with an increment of 1 each time while keeping $\lambda$$=$$0.5$ and $\gamma$$=$$0.1$. Please note that when $K$ is set as 0, our model is unable to model global interdependence between different EL decisions. Fig. \ref{KEffect-Fig} provides the experimental results. We find that our model ($K$$>$$0$) obviously outperforms its variant with $K$$=$$0$. This results indicates that graph-based evidence propagation contributes effectively to the improvement of NN-based EL method. In particular, our model achieves the best performance when $K$$=$$5$, so we set $K$$=$$5$ for all experiments.

\subsection{Overall Results}\label{Experiment_Results_Section}

Experimental results are shown in Tables \ref{in-domain} and \ref{out-domain}. We directly cited the experimental results of the state-of-the-art models reported in \cite{Francislandau:NAACL2016,Ganea:EMNLP2017,Le:ACL2018,Guo:SW2018}.
%We ran our model 3 times
%each time we picked the best model on the validation
%and then reported results on the test sets for these models.
On average, our model achieves the highest Micro F1 scores on all out-domain datasets, expect AIDA-B test set (in-domain). It is worth noting that although our model is slightly inferior to \cite{Le:ACL2018} on AIDA-B, however, its latent relations modeling can be adapted to refine our model.

\subsection{Qualitative Analysis}

\subsubsection{Effects of Recurrent Random-Walk Layers}

In table \ref{effect of random walk}, we show a hard case (where entities of two mentions are incorrectly predicted because of their low evidence scores), which is from WNED-WIKI dataset, can be correctly solved by our random-walk layers. We can notice that RRWEL($K$$=$$0$) is unable to find enough disambiguation clues from context words for the two mentions, specifically, \emph{``brown"}, \emph{ ``color"}. By contrast, RRWEL can effectively identify the correct entities with the help of the semantic interdependence based on KB: \emph{``Bay (horse)"} and \emph{``Equine coat color"} receive significantly higher evidence scores than the other candidate entities.

\subsubsection{Effects of the External KB}

In order to better understand how our model benefits from the graph-based evidence propagation based on external knowledge, we consider different kinds of transition matrix $T$ and the performances are represented in Table \ref{effect of external KB}. We can see that by leveraging external knowledge to model the semantic dependencies between candidate entities, the global interdependence between different EL decisions can be more accurately captured by our model to conduct collective EL decisions.

\section{Related Work}

In early studies, the dominant EL models are mostly statistical ones exploring manually defined discriminative features to model the local context
%\cite{Cassidy:TAC2010,Tan:COLING2010,Zheng:NAACL2010,Ji:ACL2011,Mendes:I-Semantics2011,Shen:TKDE2015}
%\cite{Ferragina:CIKM2010,Hoffart:EMNLP2011,Ratinov:ACL2011,Cheng:EMNLP2013,He:EMNLP2013,Alhelbawy:ACL2014,Ganea:WWW2015,Globerson:ACL2016,Guo:SW2018}.
compatibility \cite{Ji:ACL2011,Mendes:I-Semantics2011,Shen:TKDE2015} and the global interdependence between EL decisions \cite{Hoffart:EMNLP2011,Cheng:EMNLP2013,Ganea:WWW2015,Globerson:ACL2016,Guo:SW2018}.
With the extensive applications of deep learning in NLP, the studies of EL have marched into NN related research from resorting to conventional statistical models. Similar to the precious statistical models, the early NN-based models mainly focused on how to apply NNs to measure the local context compatibility. For example, He et al., \shortcite{He:EMNLP2013} employed Stacked Denoising Auto-encodes to learn entity representations,
%Sun et al., \shortcite{Sun:IJCAI2015} modeled mentions and entities' representations via CNNs and neural tensor networks.
Francis-Landau et al., \shortcite{Francislandau:NAACL2016} combined CNN-based representations with sparse features to model mentions and entities,
%Gupta et al., \shortcite{Gupta:EMNLP2017} presented a neural, modular system to learn entities' representations using multiple sources of information,
while Yamada et al., \shortcite{Yamada:CONLL2016} proposed a NN model to learn distributed representations for texts and KB entities.
%Recently,
%Sun et al., \shortcite{Sun:NC2018} developed a computational approach based on memory network for entity disambiguation,
%which is able to automatically find important clues of a mention from surrounding contexts,
%and leverage these clues to facilitate entity disambiguation.
Obviously, these works only focus on individual EL tasks with little attention paid on the interdependence between these target decisions. To tackle this problem,
 %Huu et al., \shortcite{Huu:COLING2016} presented Recurrent Neural Networks to adaptively compress variable length sequences of predictions for global constrains.
 Ganea et al., \shortcite{Ganea:EMNLP2017} utilized unrolled deep LBP network to model global information.
 Le and Titov \shortcite{Le:ACL2018} treated relations between textual mentions in a document as latent variables in our neural EL model.
 Moreover,
 GCN was applied by Cao et al., \shortcite{Cao:COLING2018} for the sake of integrating global coherence information for EL.
 In these works, notwithstanding semantic dependencies between various entities are able to be automatically modeled by constructing a neural network, the guidance from an external KB has always been neglected.
 %in these works, only previous EL decisions can be exploited to refine the subsequent EL ones,
 %while the interdependence between EL decisions are still far from utilization.

To address the above-mentioned issue, in this work, we exploit a neural network with recurrent random-walk layers leveraging external knowledge for collective EL. The most related works to ours include \cite{Han:SIGIR2011,Huu:COLING2016,Ganea:EMNLP2017,Guo:SW2018,Cao:COLING2018}. Significantly different from \cite{Han:SIGIR2011} and \cite{Guo:SW2018}, we resort to NN rather than statistical models for collective EL.
%As mentioned previously,
%in \cite{Huu:COLING2016}, only previous EL decisions were exploited to refine subsequent EL decisions
%while we introduce random-walk layer to fully utilize mutual effects between EL decisions.
Compared with \cite{Huu:COLING2016}, we further adopt the mutual instead of unidirectional effects between EL decisions. Also, different from \cite{Ganea:EMNLP2017,Cao:COLING2018},
%which applied Graph Convolutional Network to integrate both local features and global coherence information,
we explore external KB to model global semantic interdependence between different entities, which has been proved to be more effective in our experiments.

\section{Conclusions and Future Work}

This paper has presented a novel end-to-end neural network with recurrent random-walk layers for collective EL, which reinforce the evidence for related EL decisions into high-probability decisions with the help of external KB.
%This model jointly infers the referent entities of all mentions within the same document by exploiting both the local and the global constraints for target entity decisions.
Experimental results and in-depth analysis strongly demonstrate the effectiveness of our proposed model.

Our model is generally applicable to other tasks similar to EL,
such as word sense disambiguation, cross-lingual entity disambiguation, and lexical selection \cite{suetal2015graph}.
Therefore, we will investigate the effectiveness of our approach on these tasks.
%Since generative model has been proved to be very effective in EL \cite{Han:EMNLP2012}, we are also very interested in exploring generative NN for EL.
Besides,
%we only leverage hyperlinks of Wikipedia to model the transition matrix, and thus we would like to study how to fully exploit other resources of Wikipedia to refine our model.
we would like to fully exploit other resources of Wikipedia beyond hyperlinks to refine our model.
Finally,
inspired by the recent success of graph neural network in NLP \cite{Zhang:acl18,song2019semantic,Yin:ijcai19},
we plan to explore graph neural network based EL model in the future.

\section*{Acknowledgments}

The authors were supported by Beijing Advanced Innovation Center for Language Resources, National Natural Science Foundation of China (No. 61672440), the Fundamental Research Funds for the Central Universities (Grant No. ZK1024), Scientific Research Project of National Language Committee of China (Grant No. YB135-49), National Key Research and Development Program of China (grant No. 2016QY03D0505)
and Project S201910384407 supported by XMU Training Program of Innovation and Enterpreneurship for Undergraduates.

\bibliographystyle{named}
\bibliography{Neural_Collective_Entity_Linking_Based_on_Recurrent_Random_Walk_Network_Learning}

\end{document}